\documentclass[sigconf]{acmart}
\usepackage{graphicx}
\usepackage{multirow}
\definecolor{ForestGreen}{RGB}{34,139,34}

\usepackage{xcolor}
\AtBeginDocument{%
  }
\copyrightyear{2024}
\acmYear{2024}
\setcopyright{rightsretained}
\acmConference[CODS-COMAD Dec '24]{8th International Confernce on Data Science and Management of Data (12th ACM IKDD CODS and 30th COMAD)}{December 18--21, 2024}{Jodhpur, India}
\acmBooktitle{8th International Confernce on Data Science and Management of Data (12th ACM IKDD CODS and 30th COMAD) (CODS-COMAD Dec '24), December 18--21, 2024, Jodhpur, India}
\acmDOI{10.1145/3703323.3703338}
\acmISBN{979-8-4007-1124-4/24/12}
\begin{document}

\title{Probabilistic Trust Intervals for Out of Distribution Detection}

\author{Gagandeep Singh}
\affiliation{%
  \institution{Quansight}
  \city{Austin}
  \country{USA}
}
\email{gsingh@quansight.com}

\author{Ishan Mishra}
\affiliation{%
  \institution{Indian Institute of Technology}
  \city{Jodhpur}
  \country{India}}
\email{mishra.10@iitj.ac.in}

\author{Deepak Mishra}
\affiliation{%
  \institution{Indian Institute of Technology}
  \city{Jodhpur}
  \country{India}
}
\email{dmishra@iitj.ac.in}






\renewcommand{\shortauthors}{Singh et al.}

\begin{abstract}
  The ability of a deep learning network to distinguish between in-distribution (ID) and out-of-distribution (OOD) inputs is crucial for ensuring the reliability and trustworthiness of AI systems. Existing OOD detection methods often involve complex architectural innovations, such as ensemble models, which, while enhancing detection accuracy, significantly increase model complexity and training time. Other methods utilize surrogate samples to simulate OOD inputs, but these may not generalize well across different types of OOD data. In this paper, we propose a straightforward yet novel technique to enhance OOD detection in pre-trained networks without altering its original parameters. Our approach defines probabilistic trust intervals for each network weight, determined using in-distribution data. During inference, additional weight values are sampled, and the resulting disagreements among outputs are utilized for OOD detection. We propose a metric to quantify this disagreement and validate its effectiveness with empirical evidence. 
  Our method significantly outperforms various baseline methods across multiple OOD datasets without requiring actual or surrogate OOD samples.
  We evaluate our approach on MNIST, Fashion-MNIST, CIFAR-10, CIFAR-100 and CIFAR-10-C (a corruption-augmented version of CIFAR-10), across various neural network architectures (e.g., VGG-16, ResNet-20, DenseNet-100). On the MNIST-FashionMNIST setup, our method achieves a False Positive Rate (FPR) of 12.46\% at 95\% True Positive Rate (TPR), compared to 27.09\% achieved by the best baseline. On adversarial and corrupted datasets such as CIFAR-10-C, our proposed method easily differentiate between clean and noisy inputs. These results demonstrate the robustness of our approach in identifying corrupted and adversarial inputs, all without requiring OOD samples during training.
  
\end{abstract}




\maketitle

\section{Introduction}

Deep neural network classifiers match the human level accuracy in many applications when test data is sampled from a distribution which is same or approximately similar to the training data distribution, also known as in-distribution (ID) data \cite{intro1, vgg, resnet, intro4, intro5}. However, the networks fail in presence of data which comes from a different distribution, known as out-of-distribution (OOD) data~\cite{intro6, baseline}. These networks perform classification with an assumption of complete knowledge of the classes and therefore, wrongly classify the OOD inputs into the known categories, often with high confidence~\cite{papadopoulos2019outlier}. In real world scenarios, it is challenging to ensure the similarity of training and test data distributions \cite{ai_safety} which raises concerns on the predictions made by deep neural networks due to their vulnerability against OOD inputs. This work is motivated by the increasing demand for efficient and scalable OOD detection methods that do not require architectural changes to the network or the use of OOD surrogate data during training. Existing state-of-the-art techniques, such as deep ensembles or Bayesian networks, often provide high accuracy but at the cost of significant computational overhead or memory usage. This makes them impractical in resource-constrained environments, such as edge devices or real-time applications like autonomous systems and healthcare diagnostics, where fast and accurate detection of OOD data is critical. Our approach, based on probabilistic trust intervals, is designed to address these challenges by enabling robust OOD detection without altering the original network architecture or requiring surrogate data, providing a lightweight and scalable solution.

A solution to the above problem is to design networks which can detect OOD inputs and hence, produce reliable results. One successful design choice in this direction is to use siblings of a network. These siblings are often the copies of a single network with different weight parameter values. The siblings with similar architecture but different weights are expected to produce different outputs for the same input, therefore, are trained on ID data to produce identical outputs for an ID input. As a result, disagreement between the outputs of siblings for ID inputs reduces and it becomes an indicator of OOD inputs whenever a large value observed. For a robust detection, all the weights among the siblings are varied to explore a diverse functional space, for example in deep ensembles~\cite{ensemble, fort2019deep}. This naturally increases computational burden in terms of training and testing along with increased memory requirements. A trade-off to limit the memory expenses is to allow only few of the weights to vary among the siblings while assigning equal values to the rest or sharing the rest of the weights~\cite{wen2020batchensemble}. Another alternate to create such siblings is to learn distributions over the weight parameters, motivated from the Bayesian principles. However such networks are computationally more complex and often pose a challenge in terms of convergence for deep architectures~\cite{ensemble}.

\begin{figure}[!t]
  \centering
  \includegraphics[scale=0.045]{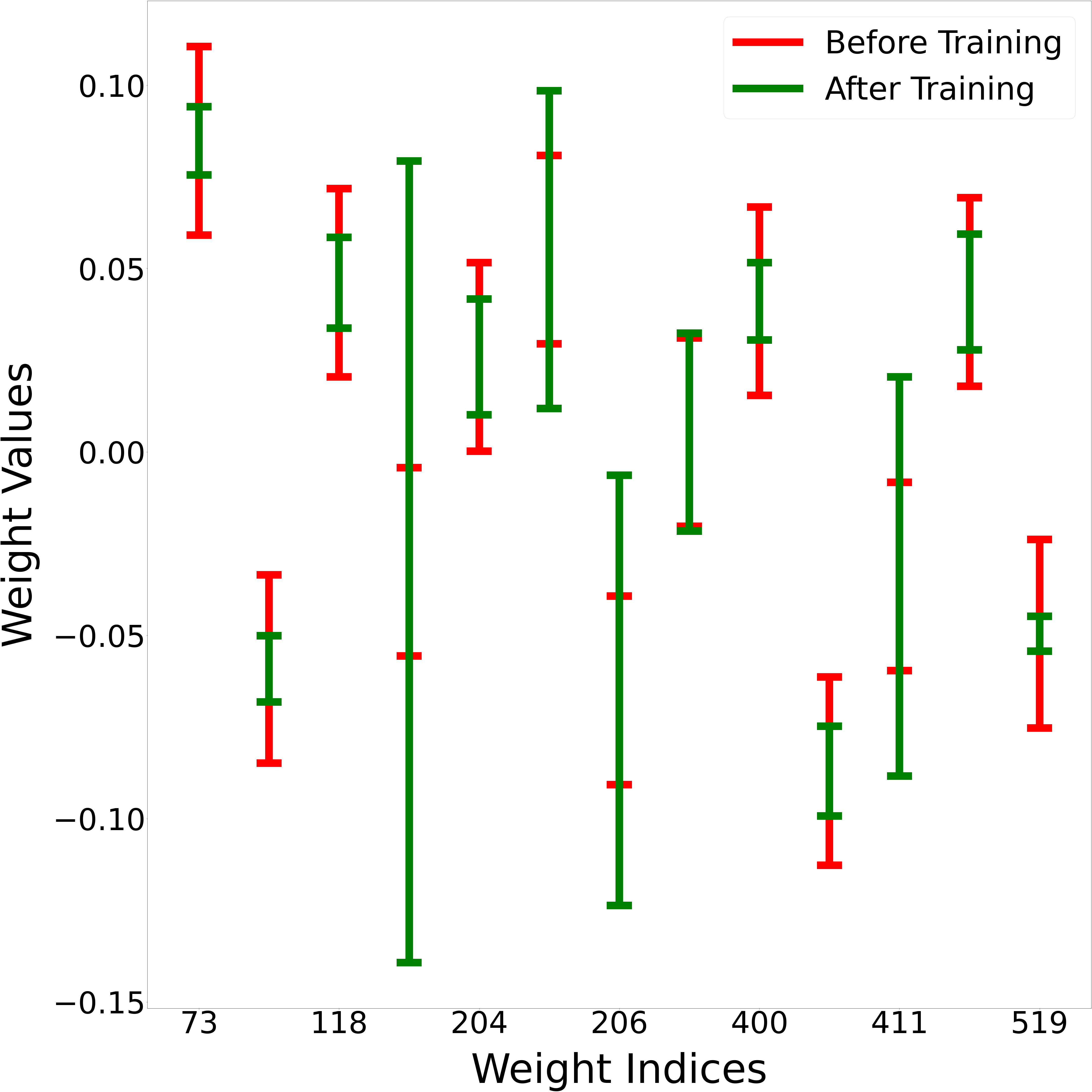}
  \Description[The image depicts the concept of Probabilistic Trust Intervals applied to the weights of a neural classifier. Two distinct states are shown: weights before training are represented in red, and weights after training are represented in green. Initially, the trust intervals are uniform across all weights, signifying equal uncertainty or variability. After training, these intervals adjust based on the in-distribution (ID) data, reflecting the learned generalization capabilities of the neural network. The adjusted intervals help in detecting out-of-distribution (OOD) inputs by indicating a deviation from the expected range of weights established during training.]{}
  \caption{Illustration of Probabilistic Trust Intervals for randomly picked weights of neural classifier before (\textcolor{red}{red}) and after (\textcolor{ForestGreen}{green}) training. Trust intervals before training are of the same size for each weight and get adjusted to ID data after training to maintain the generalization abilities of the underlying neural network while enabling OOD detection.}
  \label{fig:illus}
  \vspace{-7mm}
\end{figure}

In this paper, we propose a simple approach to create siblings of an already trained network. We define a \textit{probabilistic trust interval} for each weight around its optimized value. It allows the sampling of multiple values for each weight parameter and produces corresponding outputs for a single input. The sizes of these intervals are optimized for ID data such that the disagreement among multiple outputs for an ID input is minimized. Fig.~\ref{fig:illus} shows an illustration of probabilistic trust intervals for some of the randomly picked weights of a standard network where the observed interval sizes before (in \textcolor{red}{red}) and after (in \textcolor{ForestGreen}{green}) optimization are presented. Centers of the intervals represent the original optimized values of the corresponding weight parameters. For each parameter, values around the center are sampled with higher probability as compared to the values near to the boundary of the interval. It ensures that accuracy of the considered network on ID inputs is maintained while enabling OOD detection mechanism.



Optimization of the probabilistic trust intervals' sizes is the most critical component of our proposed approach. These intervals should be small enough to not allow the network siblings to go far off from the optimal point. At the same time, the interval sizes should be large enough to produce distinguishable output variations between ID and OOD data. For this purpose, we design a loss function which tries to optimise the size of intervals around each weight such that inside these intervals the output variations are small for ID inputs. Subsequently, these variations are used to quantify the similarity among the outputs of different sibling networks using a newly designed \textit{measure of agreement}. The deployment of probabilistic trust intervals around weights with measure of agreement on the trained networks enables OOD detection without using any kind of OOD validation samples or surrogate proxies of OOD data. 

We evaluate our method on different shallow and deep neural classifiers, including VGG-16~\cite{vgg}, ResNet-20~\cite{resnet}, and DenseNet-100~\cite{densenet20}. We use MNIST~\cite{mnist}, CIFAR-10~\cite{cifar_ten}, SVHN~\cite{svhn}, CIFAR-100~\cite{cifar_hun}, Fashion-MNIST~\cite{fmnist}, CMATERDB~\cite{cmaterdb}, and Gaussian images, where each pixel is sampled from $\mathcal{N}(0, 1)$. Our approach outperforms previous works related to OOD input detection in most of the experiments. For example, the False Positive Rate at 95\% True Positive Rate, for MNIST as ID and Fashion-MNIST as OOD data, is found to be 12.46\% using just two sets of weights, compared to 27.09\%, which is the best among the considered baselines. Additionally, we tested our approach on corrupted CIFAR-10 (CIFAR-10-C)~\cite{cifar10-c} and adversarial samples generated using the Fast Gradient Sign Method~\cite{fgsm}, observing that the proposed approach is able to comfortably highlight both corrupted and adversarial inputs.

In summary, our main contributions are as follows,
\begin{itemize}
    \item We define probabilistic trust intervals for optimized weight parameters of neural networks and provide a mechanism to optimize interval sizes according to ID data. These trust intervals allow sampling of multiple weight values to create siblings of a network.
    \item We design a measure of agreement, which in combination with optimized probabilistic trust intervals is useful to create OOD detectors.
    \item The proposed approach provides a simple way of enabling OOD detection ability in any neural network without using any OOD samples during training and also without raising any major concern of memory requirement.
    \item Our experiments show that the proposed approach is also suitable to flag corrupted and adversarial inputs.
    \item We conduct experiments with networks of various depths and with different datasets, and observe considerably better performance as compared to state-of-the-art OOD detection approaches in most cases.

\end{itemize}


\section{Related Work}
OOD detection has gained a lot of attention in recent years with an interest of creating trustable deep learning solutions. Hendrycks \& Gimpel~\cite{baseline} proposed a baseline method for OOD detection using the low softmax scores produced by a neural network for the OOD inputs. Their baseline was shown to work with a variety of datasets and neural networks on diverse sets of tasks. Subsequently, ODIN~\cite{odin} was built on top of the baseline method~\cite{baseline} which uses temperature scaling and adds small directed noise to the inputs, based on the gradient of maximum softmax scores with respect to the input. The perturbed samples can be considered as the proxies for OOD data to increase gap between the softmax scores produced by a network for ID and OOD inputs. Similar approaches where the proxies or auxiliary data is created using tools such as generative adversarial networks are reported in~\cite{lee2017training, hendrycks2018deep, papadopoulos2019outlier, thulasidasan2019mixup}. Another generative model based approach is presented in~\cite{llr}, which assumes that an input is composed of two components, a \textit{semantic} component and a \textit{background} component. Accordingly, it trains two models and proposes a log-likelihood ratio statistic on the outputs of the two models to flag OOD inputs. To train the model on background components, the dataset was created by applying directed noise on the ID samples with an assumption that it will inhibit the semantic component.
Lee et al.~\cite{mhb} also used the directed noise but in different manner where it was applied on the test inputs, which allowed the network to produce multiple outputs and select the best among them using Mahalanobis distance-based confidence scores. The approach works under the assumption that class-conditional Gaussian distributions can be fitted to the features of an already trained softmax neural classifier and subsequently can be used to measure the Mahalanobis distance of the inputs.
Recently, Fisher Information-based Evidential Deep Learning ($\mathcal{I}$-EDL)~\cite{deng2023uncertainty} has been introduced to address uncertainty estimation challenges in neural networks. Evidential neural networks~\cite{sensoy2018evidential} typically utilize a combination of expected mean squared error (MSE) and a Kullback-Leibler (KL)~\cite{kullback1951information} divergence term as a loss function, where the KL term penalizes evidence for classes not fitting the training data. $\mathcal{I}$-EDL extends this approach by incorporating the Fisher Information Matrix (FIM)~\cite{fisher1922mathematical} to measure the informativeness of evidence provided by each sample. This method allows dynamic reweighting of loss terms, focusing more on the representation learning of uncertain classes. By optimizing the PAC-Bayesian bound~\cite{mcallester1998some}, $\mathcal{I}$-EDL improves the generalization ability of the network. However, this also adds a computation overhead and complexity.

A better approach to obtain multiple outputs for a single input is to use deep ensembles~\cite{ensemble} where multiple copies of a network are trained in parallel on the ID data and variation in outputs of the ensembles during inference is used an indicator of the uncertainty in the decision and flag OOD inputs. However, this leads to a linearly increasing memory overhead with the number of ensembles used and also limits the number of outputs one can obtain. An alternate to create multiple copies of a network is BNN which learns distribution over weights and samples multiple weight values to create Bayes ensembles~\cite{blundell2015weight}. BNNs do not suffer from the problems of linearly increasing memory overhead and limited number of outputs, however, face the challenge of computational complexity and convergences of deep architectures~\cite{fort2019deep}. There have been attempts to create less complex and scalable BNNs~\cite{ritter2018scalable, deng2020bayesadapter}, however, these work on the basis of several assumptions, and often require proxies of OOD samples for uncertainty calibration and OOD detection. A PNN takes the best of both worlds in a sense that it allows sampling of multiple weight values from the trust intervals where the trust intervals can be interpreted as the space of directed noise, although added to the optimized weights of an already trained neural network, not the input samples. PNNs, therefore, do not need any proxies of the OOD samples and offer a very simple training mechanism with robust OOD detection.

Existing state-of-the-art (SOTA) methods for OOD detection, such as ODIN, Mahalanobis distance-based methods, Deep Ensembles, and Bayesian Neural Networks (BNNs), provide robust detection but often come with limitations that hinder their practical deployment. ODIN and Mahalanobis distance-based methods rely heavily on temperature scaling or additional noise injection, which increase complexity during inference. Techniques like Deep Ensembles and BNNs enhance OOD detection by creating multiple network instances or learning distributions over the network’s weights. However, these methods introduce significant computational overhead and memory usage due to the need for training and maintaining multiple models. Our proposed approach addresses these issues by offering a more efficient solution. Our method do not require modifications to the original network architecture, nor do they rely on OOD surrogate data during training. Instead, they introduce a lightweight mechanism for OOD detection by leveraging disagreements between sibling networks created through probabilistic weight sampling. This provides a scalable and computationally efficient alternative to existing methods while maintaining competitive performance.
\section{Problem Formulation and Proposed Solution}
Let $\mathbb{C}$ be a deep neural network classifier and $\mathbb{C}^{(1)}, \mathbb{C}^{(2)}, \mathbb{C}^{(3)}, \dots,\\ \mathbb{C}^{(n)}$ be its $n$ siblings. Let $P_{r}$ and $P_{e}$ be two input data distributions which are not similar, $x$ be an input sample, $M(\mathbb{C}(x), \mathbb{C}^{(1)}(x), \mathbb{C}^{(2)}(x)\\,\dots, \mathbb{C}^{(n)}(x))$ be a real-valued measure of agreement, and $\delta \in \mathbb{R}^{+}$ be a small number. The problem of OOD input detection is defined as follows:


\textit{OOD Input Detection}: If $C$ is trained on $X = \{x:x\sim P_r\}$ and receives an input $x\sim P_e$ then $M\left(\mathbb{C}(x), \mathbb{C}^{(1)}(x), \mathbb{C}^{(2)}(x),...,\mathbb{C}^{(n)}(x) \right) < \delta$ should hold. 

In what follows, we define probabilistic trust intervals which will allow the creation of siblings.

\subsection{Probabilistic Trust Intervals}

If the $i^\text{th}$ weight of an already trained neural network is denoted by $\mu_i$, and $\gamma_i \in \mathbb{R}^{+}$ is a real positive number, then the probabilistic trust interval is defined as follows,

\textit{Probabilistic Trust Interval}: $I(\mu_i) = [\mu_i - \gamma_i, \mu_i + \gamma_i]$ is the probabilistic trust interval for weight $\mu_i$ of size $2\times\gamma_i$ if the following hold
\begin{enumerate}
    \item For a weight value ($\widetilde{\mu_i}$) sampled from $I(\mu_i)$, the error in the output of the considered network ($\mathcal{E}(\widetilde{\mu_i})$) is low.
    \item The weight values near $\mu_i$ are sampled with higher probability as compared to the values away.
\end{enumerate}

\begin{figure}[!t]
\centering
\includegraphics[width=0.8\linewidth]{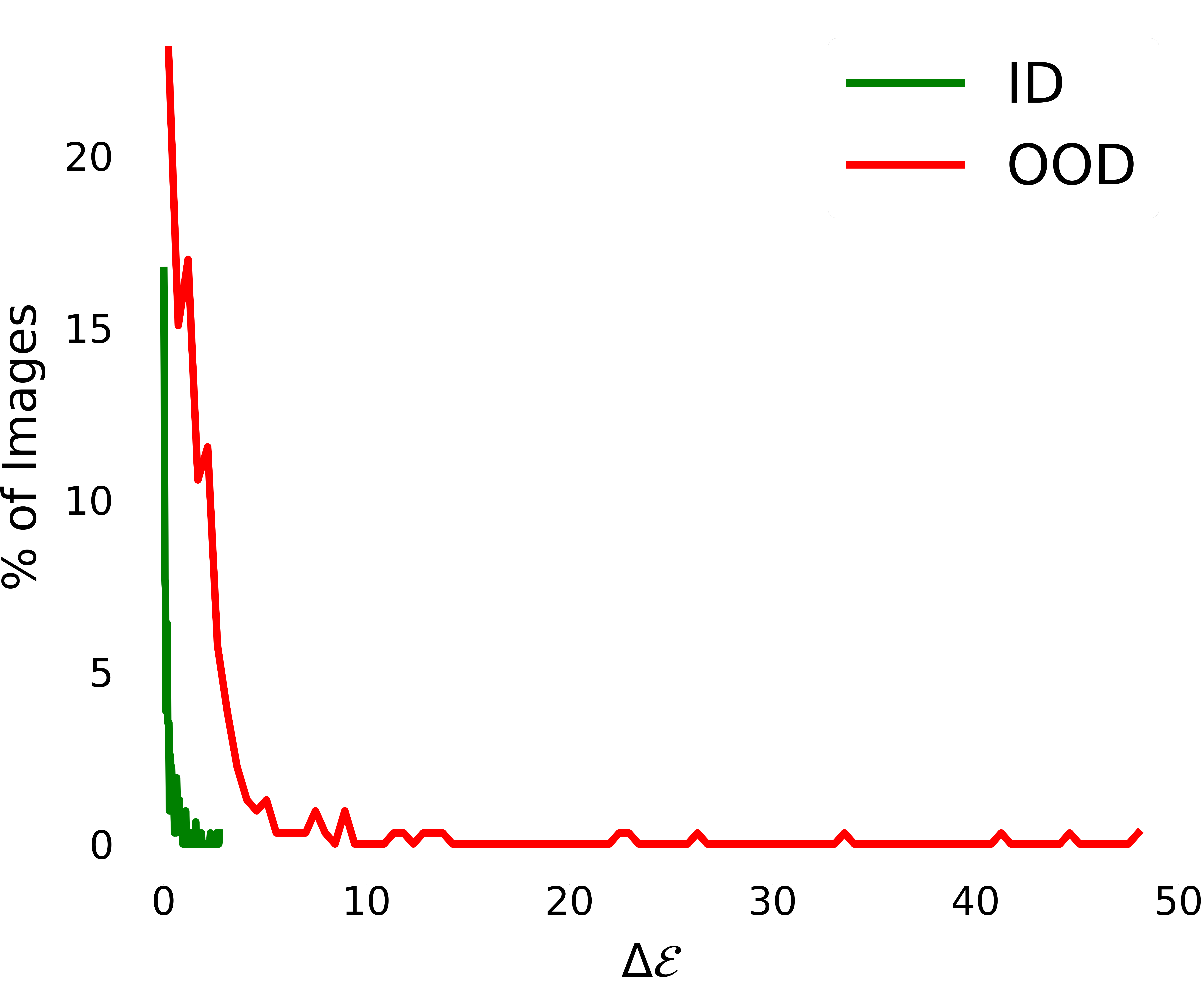}
\Description[The image illustrates the distribution of a quadratic approximation of error for the $\mathbb{C}_{1}$ neural network architecture. The MNIST dataset is used as the in-distribution (ID) data, while Fashion-MNIST is used as the out-of-distribution (OOD) data. The error values for ID samples are concentrated near zero, indicating accurate predictions and high confidence in the model's performance on familiar data. In contrast, the error values for OOD samples are distributed over a much wider range, highlighting the model's increased uncertainty and reduced accuracy when encountering data that differs from the training set. This spread in the error values demonstrates the effectiveness of the quadratic approximation in distinguishing between ID and OOD data.]{}
\caption{Distribution of quadratic approximation of error for $\mathbb{C}_{1}$ architecture with MNIST dataset as ID data and Fashion-MNIST as OOD data. It can be seen that for ID samples the values are close to zero, whereas for OOD they are spread over a considerably larger range.}
\label{fig:quad_approx}
\vspace{-2mm}
\end{figure}

In simple words, we trust that the weight values sampled from $I(\mu_i)$ will generalize well for ID data. Probabilistic nature of the interval is motivated from the training perspective, we explain this in the next section. We drop the subscript $i$ in subsequent discussion for generalization and simplicity of explanation. Let us now consider the errors in predictions from the already trained network and its sibling created using $\widetilde{\boldsymbol{\mu}} = \boldsymbol{\mu} + \Delta\boldsymbol{\mu} $ as $\mathcal{E(\boldsymbol{\mu})}$ and $\mathcal{E(\widetilde{\boldsymbol{\mu}})}$, respectively. Here $\boldsymbol{\mu}$ represents the parameter vector. As shown in~\cite{fund_calculus}, by the fundamental theorem of calculus applied on neural networks, one may find bounds on the variation in errors as:


\begin{equation}
    \Delta \mathcal{E} = \mathcal{E}(\widetilde{\boldsymbol{\mu}}) - \mathcal{E}(\boldsymbol{\mu}) \leq \max_{t} \sum_{i=1}^{L} g(\mu_{i} + t\Delta\mu_{i})^{T}\Delta\mu_{i}
    \label{ubound}
\end{equation}
Similarly,
\begin{equation}
   \min_{t} \sum_{i=1}^{L} g(\mu_{i} + t\Delta\mu_{i})^{T}\Delta\mu_{i} \leq \Delta \mathcal{E}
\end{equation}
where $g(\mu_{i})$ denotes the gradient of $\mathcal{E}(\boldsymbol{\mu})$ with respect to $\mu_{i}$ and $t \in [0, 1]$. This shows that, for a small $\Delta\boldsymbol{\mu} $, the variation in errors of siblings is bounded by the corresponding gradients. This directly gets translated to the variation in the outputs of the siblings as the ground truth used to measure the prediction errors is identical. For a network well trained on ID data, the error is expected to be minimized with low gradient values, which is not expected to be true for OOD inputs. This leads to difference in output variations which can be used for OOD detection, which is facilitated by the probabilistic trust intervals. To understand it better we consider the quadratic approximation of $\Delta \mathcal{E}$ given by the equation~\eqref{quad}
\begin{equation}
\Delta \mathcal{E} = g({\boldsymbol{\mu}})^{T}\Delta\boldsymbol{\mu} + \frac{1}{2}\Delta\boldsymbol{\mu}^{T}\mathbf{H}\Delta\boldsymbol{\mu}
\label{quad}
\end{equation}
where $\mathbf{H}$ is the Hessian matrix. Fig.~\ref{fig:quad_approx} shows the occurrences of $\Delta \mathcal{E}$ values as per equation \eqref{quad} for ID and OOD samples, where $\Delta\boldsymbol{\mu}$ is in accordance with the optimized probabilistic trust intervals of the weight parameters of a simple CNN ($\mathbb{C}_1$) reported in~\cite{cnn}. As we see, $\Delta \mathcal{E}$ values for most OOD samples is higher than the values observed for ID samples.



\subsection{Optimizing the size of Probabilistic Trust Intervals}

We model $\Delta{\mu} = \widetilde{{\mu}} - {\mu} $ as a random variable to allow sampling of multiple $\widetilde{{\mu}}$. This limits the memory expenses to a constant overhead in comparison to storing multiple values of $\widetilde{{\mu}}$ which is done in approaches like ensembles. To achieve the desired probabilistic characteristics of the trust intervals, we can consider sampling of $\widetilde{{\mu}}$ according to distributions such as Cauchy, Laplace, Triangular or any bell-shaped distribution, however we consider Gaussian distribution to take the advantage of rich literature on variational inference~\cite{variational_inference, blundell2015weight}. Accordingly, we write $\gamma = 3\times\sigma$ and $\Delta\mu =\sigma \circ \epsilon$ where $\epsilon$ is sampled from $\mathcal{N}(0, 1)$, which gives $\widetilde{\mu} = {\mu} + \sigma \circ \epsilon$.

As shown by equation~\eqref{ubound}, to ensure small variations in the outputs of siblings, it is necessary that each sibling generalizes well for the ID data and $\Delta\mu$ is small that means $\sigma$ is small. Accordingly, we define the following cost function to optimize the sizes of probabilistic trust intervals
\begin{equation}
    \mathcal{L}(D, \sigma) = \mathbb{E}_{P(\widetilde{\boldsymbol{\mu}})}\left[-\log P(D \mid \widetilde{\boldsymbol{\mu}})\right] + \pi_{1}s^2
    \label{cost1}
\end{equation}

where \(D\) represents ID data and \(\mathbb{E}\) is the expectation. The first term in the cost is the conventional negative log-likelihood of the data, which helps in retaining the generalization ability of the considered network on ID samples. The second term, \(s^{2}\), is the sum of variances of the softmax scores of each class in different outputs obtained from multiple samples of weights. \(\pi_{1}\) is a hyperparameter that controls the contribution of \(s^2\) in the total cost. Minimization of this term reduces the interval size \(\sigma\) to ensure that all siblings agree with each other on ID data for all the classes.

One obvious solution which can be produced by the cost in equation~\eqref{cost1} is $\sigma = 0$. In that case, the trust interval will collapse to a single point, and $\widetilde{\boldsymbol{\mu}} = \boldsymbol{\mu}$ will make all siblings identical to the original network. We, therefore, include $-\log(\sigma)$ as a regularizer to prevent an undesired collapse and allow the discovery of a non-zero value of $\sigma$ to facilitate OOD detection. Other suitable regularizers such as negative $\ell_1$ or $\ell_2$ can also be used. The updated cost function is as follows:
\begin{equation}
    \mathcal{L}(D, \sigma) = \mathbb{E}_{P(\widetilde{\boldsymbol{\mu}})}\left[-\log P(D \mid \widetilde{\boldsymbol{\mu}})\right] + \pi_{1}s^2 - \pi_{2} \log(\sigma)
    \label{cost2}
\end{equation}
where \(\pi_{2}\) controls the effect of the regularizer. In short, $\mathcal{L}$ in equation~\eqref{cost2} ensures that the probabilistic trust interval for each weight will be non-zero in size and will be optimized in such a way that the disagreement between multiple samples of $\widetilde{\boldsymbol{\mu}}$ is minimized for the given data without compromising on the test accuracy.

\begin{figure}
\centering
   \includegraphics[width = \linewidth]{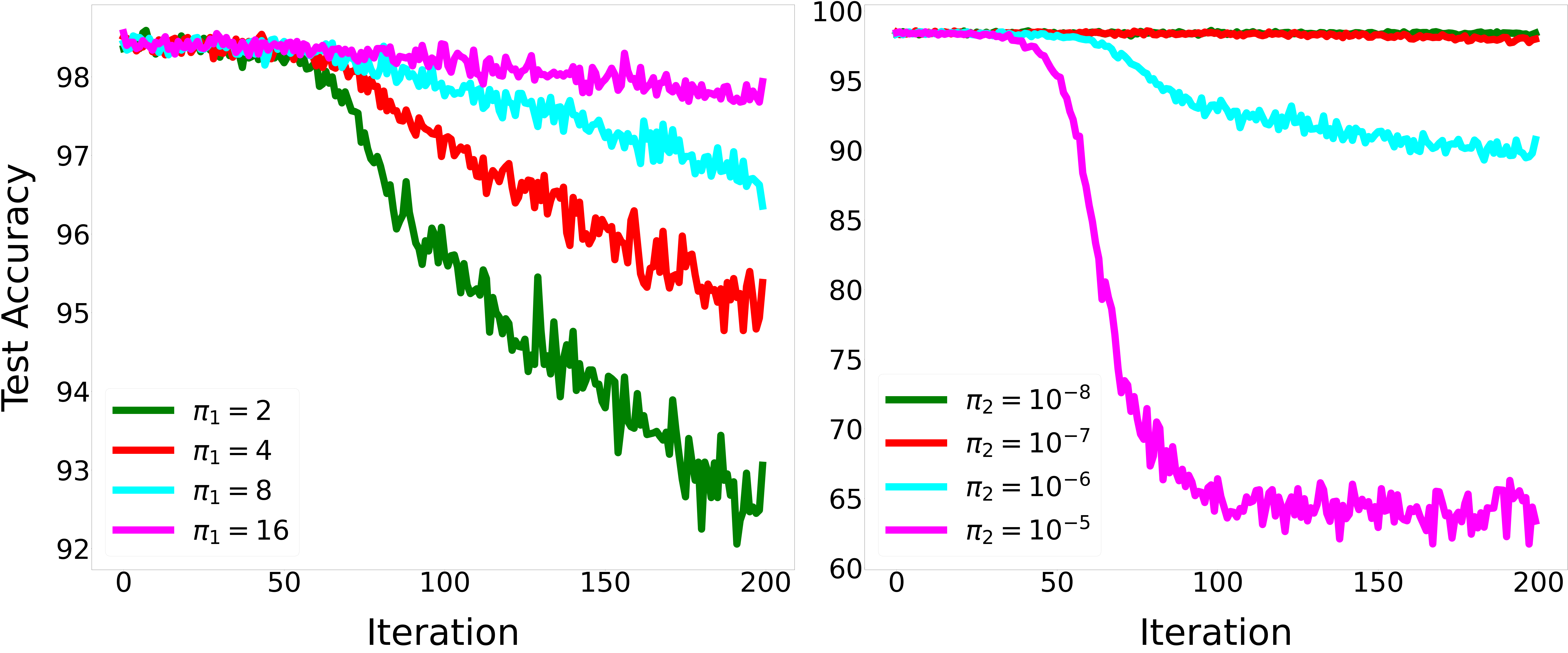}
\Description[The image consists of two plots illustrating the change in test accuracy as a function of updates in a neural network model. In the left plot, the parameter $\pi_{1}$ is varied across different values while keeping the parameter $\pi_{2}$ constant at $10^{-6}$. This plot shows how changes in $\pi_{1}$ impact the model's accuracy over time. In the right plot, the parameter $\pi_{2}$ is varied while $\pi_{1}$ is held fixed at 1, demonstrating the effect of different $\pi_{2}$ values on the accuracy.]{}
\caption{The change in test accuracy with updates. In the left plot different values of $\pi_{1}$ is varied while $\pi_{2}$ is fixed to $10^{-6}$ and in the right plot different values of $\pi_{2}$ is taken while $\pi_{1}$ is fixed to 1.} 
\label{fig:train_cost_plots}
\vspace{-2mm}
\end{figure}

It is interesting to analyze the effect of $\pi_{1}$ and $\pi_{2}$ on the trust interval size and test accuracy. For this purpose, we consider a simple CNN ($C_{1}$) reported in~\cite{cnn} with MNIST as ID data to observe the variation in test accuracy of the resultant siblings. These observations are plotted in Fig.~ \ref{fig:train_cost_plots}. As can be seen small values of $\pi_{1}$ are unable to maintain the test accuracy of the resultant siblings. On the other hand for small values of $\pi_{2}$ the test accuracy remains consistent, however, as we increase the value of $\pi_{2}$, the generalization capabilities of the network drop with increase in $\sigma$ and stabilizes towards the end. Accordingly a trade-off between the values of $\pi_{1}$ and $\pi_{2}$ is used in the experiments. The proposed approach provides a network where sampling of multiple weight values from the probabilistic trust interval effectively results in controlled perturbations on the optimized weights with certain characteristics, therefore, we call this new network as Perturbed-NN or PNN. Next, we define a suitable measure of agreement to quantify the variation in outputs of a single input and use it for OOD detection.


\subsection{Measure of Agreement}

PNN allows us to obtain multiple outputs for a single input by sampling weights using $\mu$ and $\sigma$. Now, we need a measure which can help in effectively quantifying the variations in the outputs of the PNN for flagging OOD inputs. As opposed to many of the previous works, we consider both average value and standard deviation of the outputs of PNN. Formally, suppose that \(n\) Monte Carlo samples of weights are drawn using PNN's parameters for classifying the input to \(c\) classes, thereby giving us \(n\) softmax scores for each of the \(c\) classes. Further suppose that, the mean softmax score and standard deviation for \(k^{th}\) class is \(\alpha_{k}\) and \(\beta_{k}\) (slightly deviating from the standard notations for mean i.e., $\mu$ and standard deviation i.e., $\sigma$, for avoiding confusion) respectively. We can see that \(\frac{\beta_{k}^{2}}{\alpha_{k}}\) is nothing but the index of dispersion of \(n\) softmax scores for the \(k^{th}\) class. Thus, the quantity $\sum_{k=1}^{c} \frac{\beta_{k}^{2}}{\alpha_{k}}$ will attain a low value only if the output variation and index of dispersion for all the classes is low. This quantity can now be combined with entropy of the average softmax scores of the classes to provide the desired measure of agreement as,

\begin{equation}
    M =\frac{1}{\sum_{k=1}^{c} \frac{\beta_{k}^{2}}{\alpha_{k}}} + \frac{1}{\sum_{k=1}^{c} -\alpha_{k}\log{\alpha_{k}}}
     \label{measure}
\end{equation}
A large value of $M$ indicates high level of agreement between the outputs of different siblings. A small constant in the denominator of $M$ can be added for stability.

In order to explain why $M$ is better than other measures used in previous works, we have made a conceptual comparison in the following points.

\textit{Shannon's Entropy} - Usually, the entropy of the average softmax scores obtained from different siblings, including ensembles and Bayesian neural network (BNN), is used as a measure of uncertainty to identify OOD inputs. One problem with this is that by using only the average softmax scores, the variances are ignored. Hence, through the entropy values, only a partial knowledge of the amount of agreement between the multiple sets of weights for a given input is captured. In contrast, $M$ in equation~\eqref{measure} takes into account the variances of the output scores along with the entropy of average scores to extract more information.

\textit{Maximum of average softmax scores} - This measure has been used in \cite{baseline, odin} to detect OOD inputs. Similar to the entropy of average scores, the spread in average scores for different classes is ignored if we only use the maximum of the averages. Our measures give more weight to the variances of the classes with lower softmax scores (observe the $\frac{1}{\alpha_{k}}$ multiplied with $\beta_{k}^{2}$ in equation~\eqref{measure}) as compared to those with higher softmax scores. This means if a sibling is selecting a particular class with higher confidence, then it should reject other classes in agreement with the other siblings. Otherwise, the index of dispersion for the rejected classes will increase, resulting in PNN suspecting the input as OOD.

\textit{Sum of KL divergence between individual and the average softmax scores} - This measure was used by \cite{ensemble} for quantifying uncertainty in the outputs of an ensemble of neural networks. A conceptual limitation of this measure arises in the scenario when the same class is accepted with high softmax scores by all the networks in the ensemble. In this case, the KL divergence between the individual and the average softmax scores will be dominated by the class with maximum softmax score. Hence, the agreement between the networks with respect to all the classes will not be reflected entirely. 
\section{Experiments and Results}
We use the following combination of architectures and ID datasets in our experiments and corresponding PNNs.

${\mathbb{C}_{1}}$ - A shallow CNN architecture containing only two convolutional and two fully connected layers~\cite{cnn}. The first two layers are convolution layers with $5\times 5$ kernels and number of channels as 32 and 64, respectively. Each convolutional layer output is downsampled using max pooling, and the resultant output passes through the two fully connected layers containing 1024 and 10 output units, respectively, to produce the final output. All but the last layer uses ReLU activation. MNIST dataset is used as ID data. We train the corresponding PNN for 1708 iterations with a batch size of 256. 

${\mathbb{C}_{2}}$ - The second architecture is VGG-16 for which we used CIFAR-10 as ID data to create the corresponding PNN in 1712 iterations with a batch size of 64.

${\mathbb{C}_{3}}$ - Third architecture considered is ResNet-20 on CIFAR-10 dataset. The corresponding PNN converges in 802 iterations with a batch size of 128.

${\mathbb{C}_{4}}$ - This is also a simple CNN architecture trained on SVHN dataset. It contains a sequence of convolution-batch-normalization-convolution-max-pool-dropout blocks. There are three such blocks with 32, 64 and 128 channels in the convolutional layers. The kernel size of all the convolutional layers is $3\times 3$. The window size for max-pool is $2\times 2$. The dropout rate is 0.3. The last two layers are fully connected layers with 128 and 10 output units. For PNN, we use 315 iterations with a batch size of 128.

${\mathbb{C}_{5}}$ - Final architecture is a comparatively deeper architecture - DenseNet-100 with a growth rate of 12. It is trained on CIFAR-10 dataset. For PNN model of this architecture, we use 120 iterations with a batch size of 64.

The performance of our proposed method is evaluated using widely accepted OOD detection metrics:
\begin{itemize}
    \item False Positive Rate (FPR) at 95\% True Positive Rate (TPR): This metric measures the rate of false alarms when the system correctly identifies 95\% of ID inputs. A lower FPR is desirable for reliable OOD detection.
    \item Area Under the Precision-Recall Curve (AUPR): AUPR provides a threshold-independent performance evaluation, especially useful in scenarios where OOD inputs are rare. A higher AUPR indicates better overall precision and recall for detecting OOD inputs.
\end{itemize}

In addition to these standard metrics, we introduce a measure of agreement specifically designed for our methodology. This measure quantifies the level of disagreement between outputs of sibling networks (created by sampling from the probabilistic trust intervals). The idea is that ID inputs will yield similar outputs across the sibling networks, while OOD inputs will result in more varied outputs. The measure of agreement is calculated using both the mean softmax scores and the standard deviation of the outputs. A lower value of this measure indicates high agreement and suggests the input is ID, while a higher value indicates low agreement and flags the input as OOD.

By combining these standard metrics with the measure of agreement, we provide a comprehensive evaluation of our method’s effectiveness in distinguishing ID from OOD data across multiple benchmarks.


To ensure numerical stability during minimization of the cost in equation~\eqref{cost2}, we parameterize \(\sigma \) using \(\rho\) as, $\sigma = \log(1 + \exp(\rho))$, and $\rho$ is initialized using $U(0, 1)$.
The optimizer used for all the architectures is \(RMSprop\)~\cite{rmsprop} with a learning rate of 0.01. 
The values of \(\pi_{2}\) are kept less than $10^{-6}$, which allows \(\pi_{1}\) to be fixed at 1 for all of our experiments without raising any concerns of convergence. We stopped the training when either the negative log likelihood, cross entropy for image classification tasks, started to increase or there wasn't any further decrease in \(\pi_{1}s^2\). The intuition behind this is that optimizing \(\sigma\) in further iterations won't be beneficial because, increasing \(\sigma\) will either fluctuate the outputs for ID data too much and decreasing \(\sigma\) will increase \(-\log (\sigma)\). 

To begin with, we first evaluate the obtained PNNs performances on ID data. Table~\ref{tab:acc} shows the differences in mean classification accuracy of CNNs with point estimate and the corresponding PNNs on ID test samples. For each PNN only a single sample of weights is drawn from their trust intervals for computing the test accuracy. It can be observed that the change in test accuracy is negligible. This shows that when an already trained neural classifier is converted into a PNN using the proposed approach, its performance on ID test data is maintained. Next we show that while maintaining the performance on ID data, PNNs facilitate the detection of OOD inputs.

\begin{table}
\caption{Test accuracy of CNNs with point estimate and the corresponding PNNs for various architectures. For all the architectures, PNN accuracy is very close to CNN which shows that PNNs maintain the performance of corresponding CNNs on ID data while enabling OOD detection.}
\label{tab:acc}
\begin{center}
\begin{tabular}{cccc}
\toprule
\bf Architecture & \bf Dataset & \bf CNN (in \%) & \bf PNN (in \%) \\
\midrule 
${\mathbb{C}_{1}}$(ShallowCNN) & MNIST & 98.53 & 98.49 \\
${\mathbb{C}_{2}}$(VGG-16) & CIFAR10 & 93.54 & 93.66 \\
${\mathbb{C}_{3}}$(ResNet20) & CIFAR10 & 91.68 & 91.44 \\
${\mathbb{C}_{4}}$(SimpleCNN) & SVHN & 95.93 & 94.43 \\
${\mathbb{C}_{5}}$(DenseNet-100) & CIFAR10 & 93.51 & 93.55 \\
\midrule 
\end{tabular}
\end{center}

\end{table}
\begin{table*}
\caption{Comparison of our proposed approach with the state-of-the-art methods with respect to FPR at 95\% TPR. Low FPR at high TPR is important when it comes to selecting a threshold for deployment of an OOD detection technique}
\label{tab:fpr}
    \centering
    \begin{tabular}{ccccccccccc}
      \toprule 
      \bfseries Model & \bfseries ID & \bfseries OOD & \bfseries Baseline & \bfseries ODIN & \bfseries BayesAdapter & \bfseries DeepEnsemble & \bfseries MHB & \bfseries $\mathcal{I}$-EDL & \bfseries PNN (ours) \\
      \midrule 
      \bfseries ${\mathbb{C}_{1}}$ & MNIST & FMNIST & 27.48 & 27.09 & 70.61 & 36.86 & 30.31 & 28.84 & \textbf{12.46}\\
      \midrule 
      \multirow{2}{*}{\bfseries ${\mathbb{C}_{2}}$} & CIFAR10 & CMATERDB & 61.48 & 56.80 & 58.63 & 88.32 & 100.0 & 62.24 & \bfseries 54.80\\
      & CIFAR10 & CIFAR100 & 71.35 & 71.39 & 70.15 & 71.46 & 99.86 & 69.72 & \textbf{67.41}\\
      \midrule 
      \multirow{2}{*}{\bfseries ${\mathbb{C}_{3}}$} & CIFAR10 & CMATERDB & 58.88 & 59.03 & 66.11 &  44.12 & 75.63 & \bfseries 42.25 & 48.18\\
      & CIFAR10 & CIFAR100 & 70.74 & 71.6 & 76.73 & 82.97 & 99.71 & 68.34 & \textbf{64.37}\\
      \midrule 
      \multirow{3}{*}{\bfseries ${\mathbb{C}_{4}}$} & SVHN & CIFAR10 & 43.80 & 43.18 & 71.00 & 66.12 & 88.50 & 45.85 & \bfseries 37.00\\
      & SVHN & CMATERDB & 49.00 & 52.59 & 73.45 & \bfseries 45.54 & 87.08 & 46.75 & 51.00\\
      & SVHN & GAUSSIAN & 19.95 & 19.97 & 45.36 & 60.73 & 95.08 & 22.5 & \bfseries 1.61\\
      \midrule 
       \bfseries ${\mathbb{C}_{5}}$ & CIFAR10 & CIFAR100 & 70.18 & 69.71 & 72.37 & 69.71 & 97.122 & 71.42 & \bfseries 68.71\\
      \bottomrule 
    \end{tabular}

\end{table*}

\begin{table*}
\centering
\caption{Comparison of our proposed approach with the state-of-the-art techniques with respect to AUPR. The proposed PNNs result in comparatively better performance in most experiments}
\label{tab:aupr}
    \begin{tabular}{ccccccccccc}
      \toprule 
      \bfseries Model & \bfseries ID & \bfseries OOD & \bfseries Baseline & \bfseries ODIN & \bfseries BayesAdapter & \bfseries DeepEnsemble & \bfseries MHB & \bfseries $\mathcal{I}$-EDL & \bfseries PNN (ours) \\
      \midrule 
      \bfseries ${\mathbb{C}_{1}}$ & MNIST & FMNIST & 97.00 & 96.92 & 51.60 & 55.72 & 96.49 & 98.76 & \bfseries 99.44\\
      \midrule 
      \multirow{2}{*}{\bfseries ${\mathbb{C}_{2}}$} & CIFAR10 & CMATERDB & 92.58 & 93.11 & 78.21 & 60.00 & 62.01 & 93.5 & \bfseries 95.98\\
      & CIFAR10 & CIFAR100 & 82.56 & 82.88 & 68.51 & 57.56 & 40.89 & 85.4 & \bfseries 88.92\\
      \midrule 
      \multirow{2}{*}{\bfseries ${\mathbb{C}_{3}}$} & CIFAR10 & CMATERDB & 90.68 & 90.10 & 88.27 & 55.22 & 90.60 & 91.2 & \bfseries 92.41\\
      & CIFAR10 & CIFAR100 & 85.89 & 85.63 & 82.88  & 66.49 & 50.74 & 88.5 & \bfseries 98.82\\
      \midrule 
      \multirow{3}{*}{\bfseries ${\mathbb{C}_{4}}$} & SVHN & CIFAR10 & 93.12 & 93.02 & \bfseries 96.99 & 56.59 & 92.70 & 93.7 & 94.85\\
      & SVHN & CMATERDB & 84.93 & 84.49 & \bfseries 93.64 & 60.60 & 87.68 & 85.7 & 87.73\\
      & SVHN & GAUSSIAN & 98.00 & 97.92 & 96.91 & 54.78 & 91.41 & 97.4 & \bfseries 98.51\\
      \midrule 
       \bfseries ${\mathbb{C}_{5}}$ & CIFAR10 & CIFAR100 & 69.59 & 69.93 & 66.05 & 64.03 & 70.56 & 75.6 & \bfseries 78.3\\
      \bottomrule 
    \end{tabular}

\end{table*}

We consider different combinations of ID and OOD datasets for different architectures. We compare OOD detection performance of PNN with existing works including state-of-the-art approaches such as Deep Ensembles~\cite{ensemble},  BayesAdapter~\cite{deng2020bayesadapter} and $\mathcal{I}$-EDL~\cite{deng2023uncertainty} along with ODIN~\cite{odin}, MHB~\cite{mhb}, and the Baseline OOD detection approach~\cite{baseline}. For ODIN, we pick \(T\) for temperature scaling from, {10, 100, 1000} and \(\epsilon\) from \{0.0001, 0.00625, 0.025, 0.05, 0.1\}. In Mahalanobis distance based OOD detection (MHB), we use the features from the penultimate layer and the layer preceding it. Similarly, for Deep Ensembles, BayesAdapter, and the proposed PNNs, we considered two siblings of each for a fair comparison. For $\mathcal{I}$-EDL, we follow the author's code for hyper-paramters. It should be noted that, no method including ours, is adapted to OOD data or calibrated for uncertainty using proxies of OOD data, as in a general scenario, the OOD examples can come from any distribution which may or may not be known during training. 
We use False Positive Rate at 95 \% True Positive Rate (FPR at 95 \% TPR), Area Under the Precision Recall Curve (AUPR), and Area Under the Receiver Operating Characteristic Curve (AUROC) as the performance metrics. The results for FPR at 95 \% TPR and AUPR are shown in Tables~\ref{tab:fpr} and \ref{tab:aupr}, respectively. AUROC results are provided in the supplementary material. It can be seen that, PNNs outperform the previous works in most of the experiments and achieve comparable results in other, particularly in terms of AUPR. It is noteworthy that since AUPR is a threshold independent measure it can be of great significance when it comes to comparing different techniques for OOD detection. Higher values of AUPR reflect better performance on a large number of thresholds. Our method introduces probabilistic trust intervals around the network’s weights, which allows for controlled perturbations during inference. Unlike baseline methods like Deep Ensembles, which require training multiple independent models, or BNNs, which need to learn distributions over all weights, your approach adapts the perturbation around existing weights without modifying the original network architecture. This provides an efficient way to explore variations in the network’s decision space, leading to better uncertainty estimation.
The adaptive flexibility provided by trust intervals ensures that the network can maintain performance on ID data (i.e., low variance in sibling outputs) while allowing for greater variability in outputs when OOD data is encountered. This balance between stability and flexibility is something that other baseline methods struggle with, as they often require heavy computation or fail to generalize well across different datasets and architectures. Many baseline methods, such as ODIN or Mahalanobis-based OOD detection, either implicitly or explicitly rely on surrogate OOD samples or assumptions about data distributions (e.g., class-conditional Gaussian distributions). These assumptions and dependencies can break down when the OOD data does not match the surrogate data or violates the distributional assumptions. To understand the capacity of the proposed PNNs better, we further test them against corrupted and adversarial inputs, and present the results as follows.

\subsection{Experiments with Noisy Test Samples}

\begin{figure}[!t]
\begin{center}
  \includegraphics[width = \linewidth]{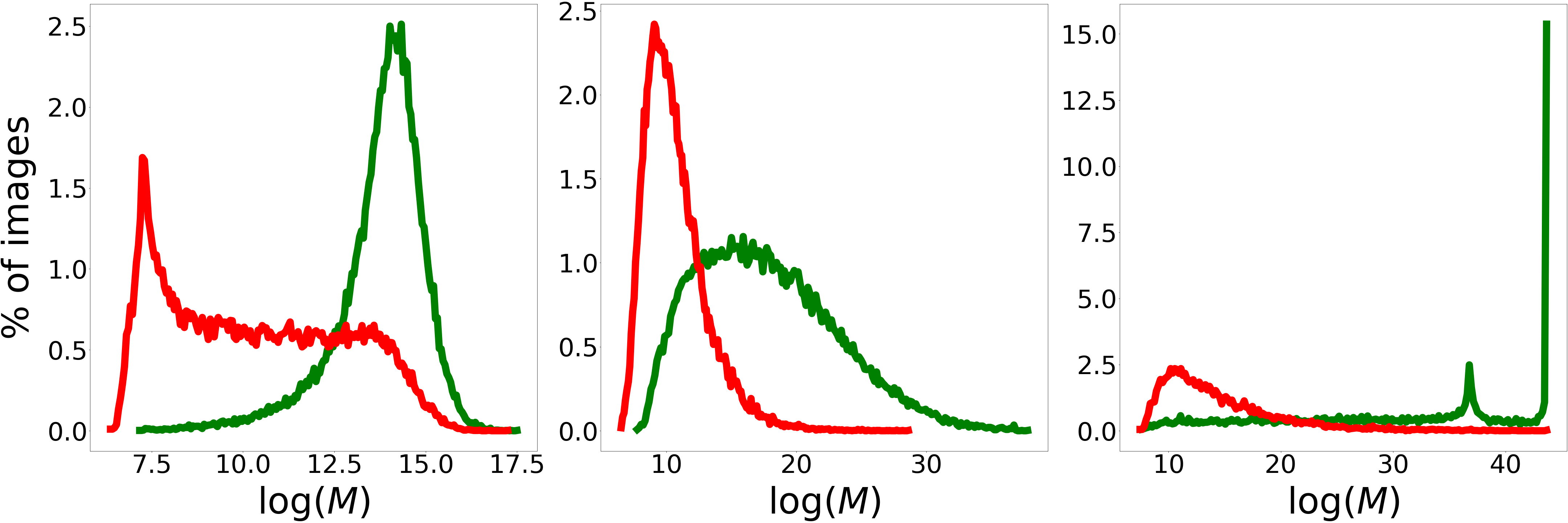}
  
  \vspace{0.2cm}
  \includegraphics[width = \linewidth]{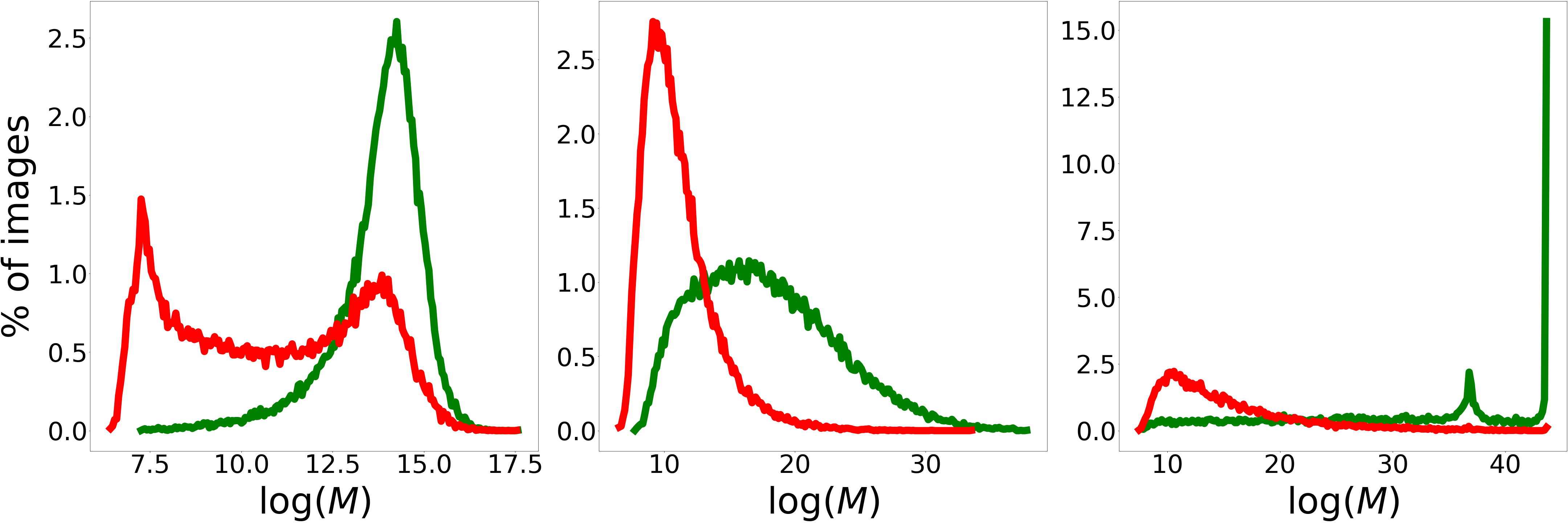}
\end{center}
\vspace{-0.2cm}
\Description[The image shows two sets of distributions comparing the values of $\log{(M)}$ for original clean images and their corrupted versions from the CIFAR-10-C dataset. The clean images are represented in green, while the corrupted images are shown in red. The top row displays the distribution for images corrupted with Gaussian noise, and the bottom row shows the distribution for images corrupted with Speckle noise. Each plot highlights the distinct differences in the distributions between clean and noisy samples.]{}
\caption{Distribution of values of $\log{(M)}$ for the original clean images (in \textcolor{ForestGreen}{green}) and their Gaussian noise corrupted versions (top row, in \textcolor{red}{red}) and Speckle noise corrupted versions (bottom row, in \textcolor{red}{red}) obtained from CIFAR-10-C dataset. A distinct nature of the plots for clean and noisy samples can be observed.}
\label{fig:cifar10c}
\end{figure}

We use noisy versions of ID CIFAR-10 samples, known as CIFAR-10-C~\cite{hendrycks2019robustness}. We consider two of the most commonly observed real world noises, Gaussian and Speckle, and pass them to PNNs corresponding to ${\mathbb{C}_{2}}$, ${\mathbb{C}_{3}}$ and ${\mathbb{C}_{5}}$ for which CIFAR-10 is used as the ID. We plot distribution of the observed values of the measure of agreement ($M$) on a logarithmic scale in Fig.~\ref{fig:cifar10c}. The figure also shows distribution of the values of $M$ observed for the clean ID samples. A clear difference can be observed for noisy and clean samples, particularly for the deep architecture DenseNet-100. Notice the consistency in the observed distribution of the values of $M$ for clean samples between the plots in the top and bottom row of Fig.~\ref{fig:cifar10c} for different architectures. Even though different values are sampled for the two different types of noisy samples, a negligible variation is observed while using these weights for clean samples. This shows the robust behaviour of PNNs.



\subsection{Experiments with Adversarial Test Samples}
\begin{figure}
\begin{center}
   \includegraphics[width = \linewidth]{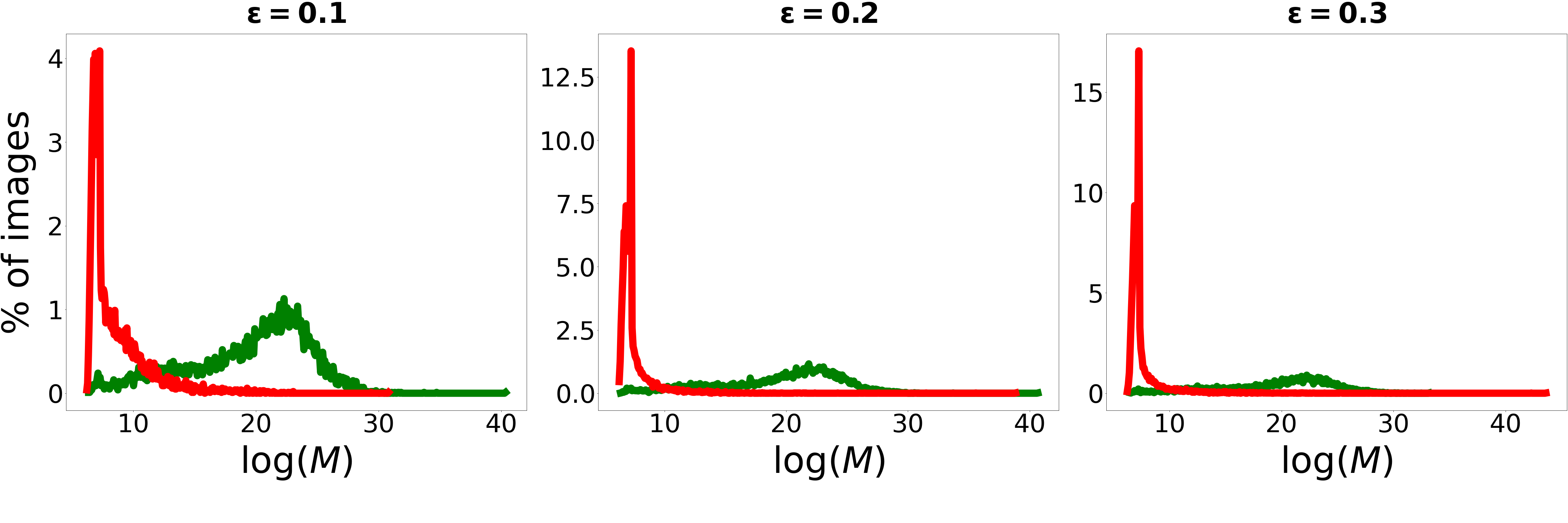}
   \includegraphics[width = \linewidth]{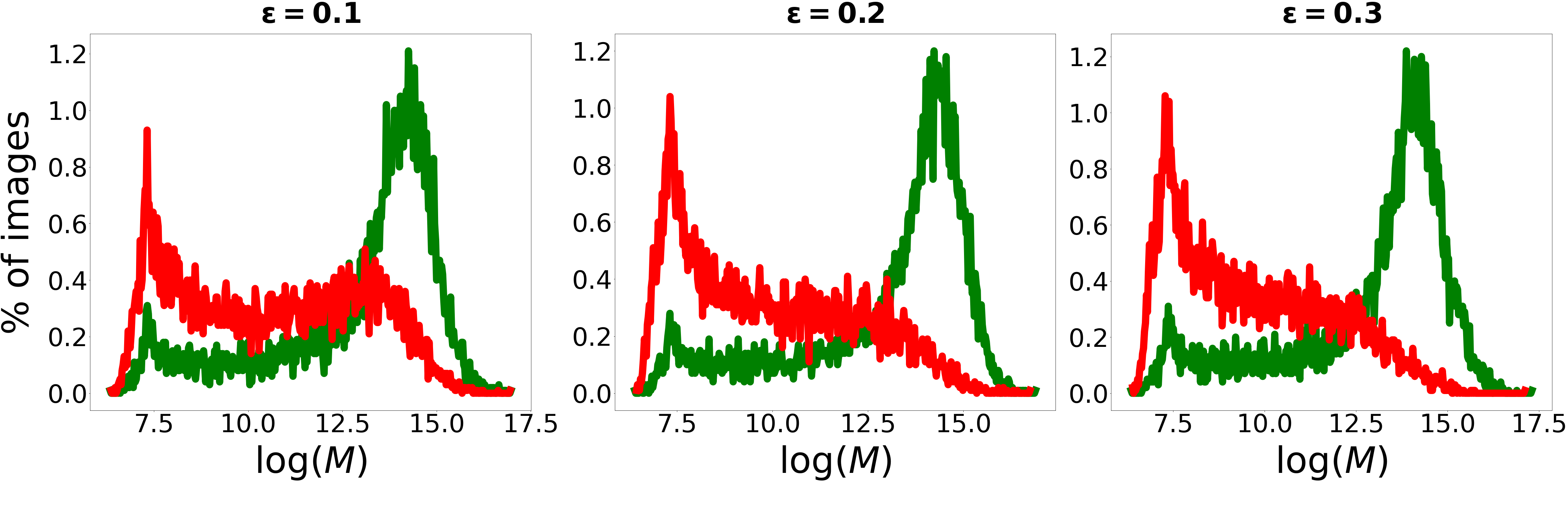}
   \includegraphics[width = \linewidth]{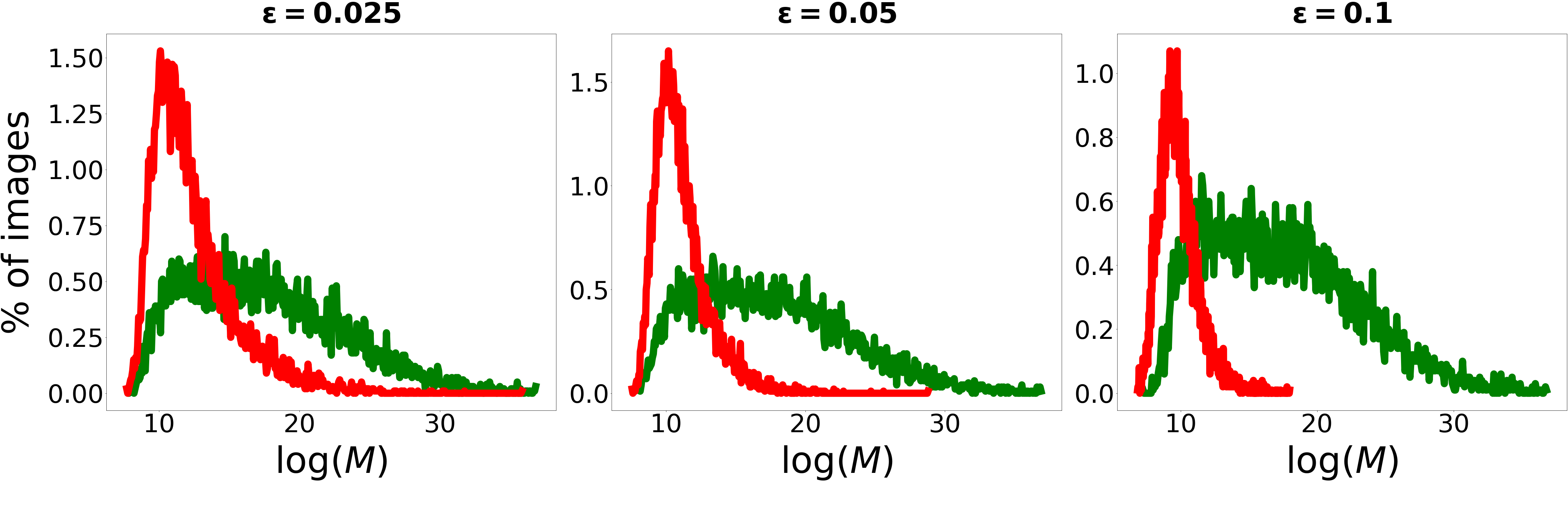}
   \includegraphics[width = \linewidth]{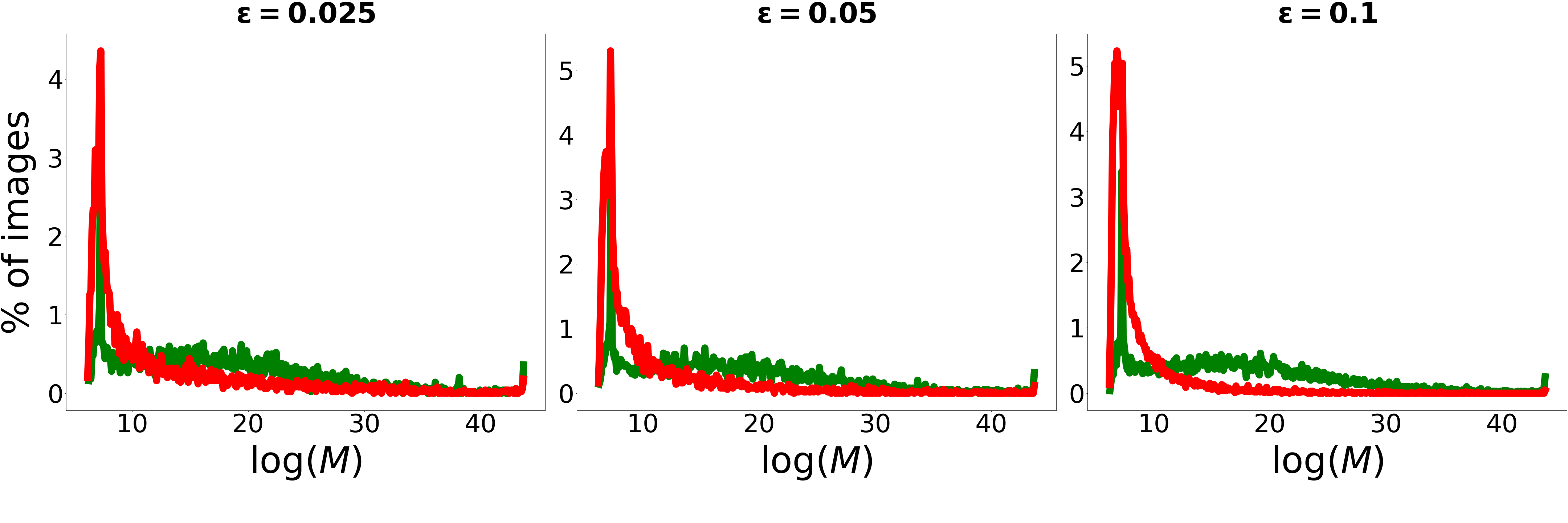}
   \includegraphics[width = \linewidth]{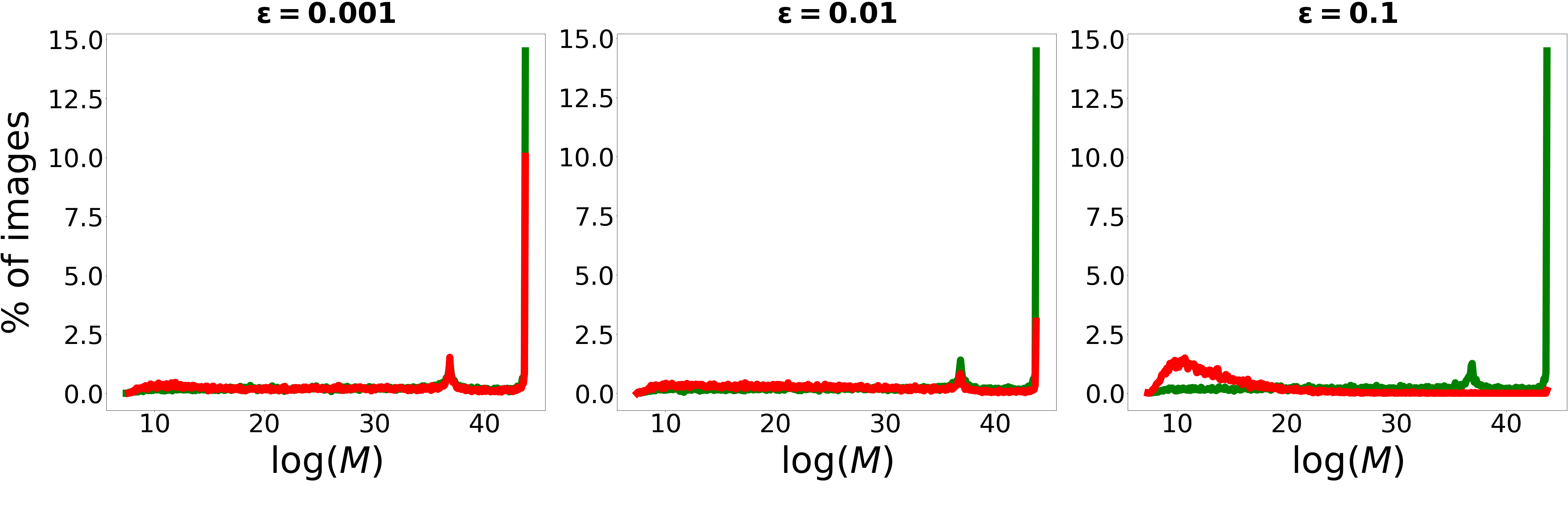}
\end{center}
\Description[The image presents the distribution of $\log{(M)}$ for original images, shown in green, and their adversarial versions, shown in red. The rows from top to bottom correspond to observations for models ${\mathbb{C}{1}}$ through ${\mathbb{C}{5}}$, respectively. The peaks of the distributions for the original and adversarial images are distinctly separated, indicating clear differences. Additionally, as the intensity of the adversarial attack, denoted by $\epsilon$, increases, the overlap between the distributions diminishes. A noticeable shift in the red peak occurs with increasing $\epsilon$, highlighting the escalating impact of the adversarial attacks on the images.]{}
\caption{Distribution of $\log{(M)}$ for original images (in \textcolor{ForestGreen}{green}) and their adversarial versions (in \textcolor{red}{red}). Rows from top to bottom show the observations for ${\mathbb{C}_{1}}$ to ${\mathbb{C}_{5}}$, respectively. Peeks of the distribution for original and adversarial images can be easily distinguished from each other. Moreover, with the increase in intensity of adversarial attack ($\epsilon$), the overlap between the distributions decreases. Observe the shift in the red peak with increasing $\epsilon$.}
\label{fig:adv_plots_m2}
\vspace{-3mm}
\end{figure}

Next we consider adversarial images generated from the original ID samples using FGSM attack~\cite{fgsm}. FGSM is applied on each architecture with its point estimates and adversarial samples are generated for the corresponding ID data which was used during training for obtaining those point estimates. Similar to the previous experiment, distributions of the observed values of $M$ for the original and adversarial samples are plotted in Fig.~\ref{fig:adv_plots_m2}. As the intensity of the attack increases, the quality of the samples degrades, and PNNs highlight it by bringing out clear separations in the peaks of the distribution corresponding to the original and adversarial samples.

\begin{figure}[!t]
\centering
  \includegraphics[width = 0.6\linewidth]{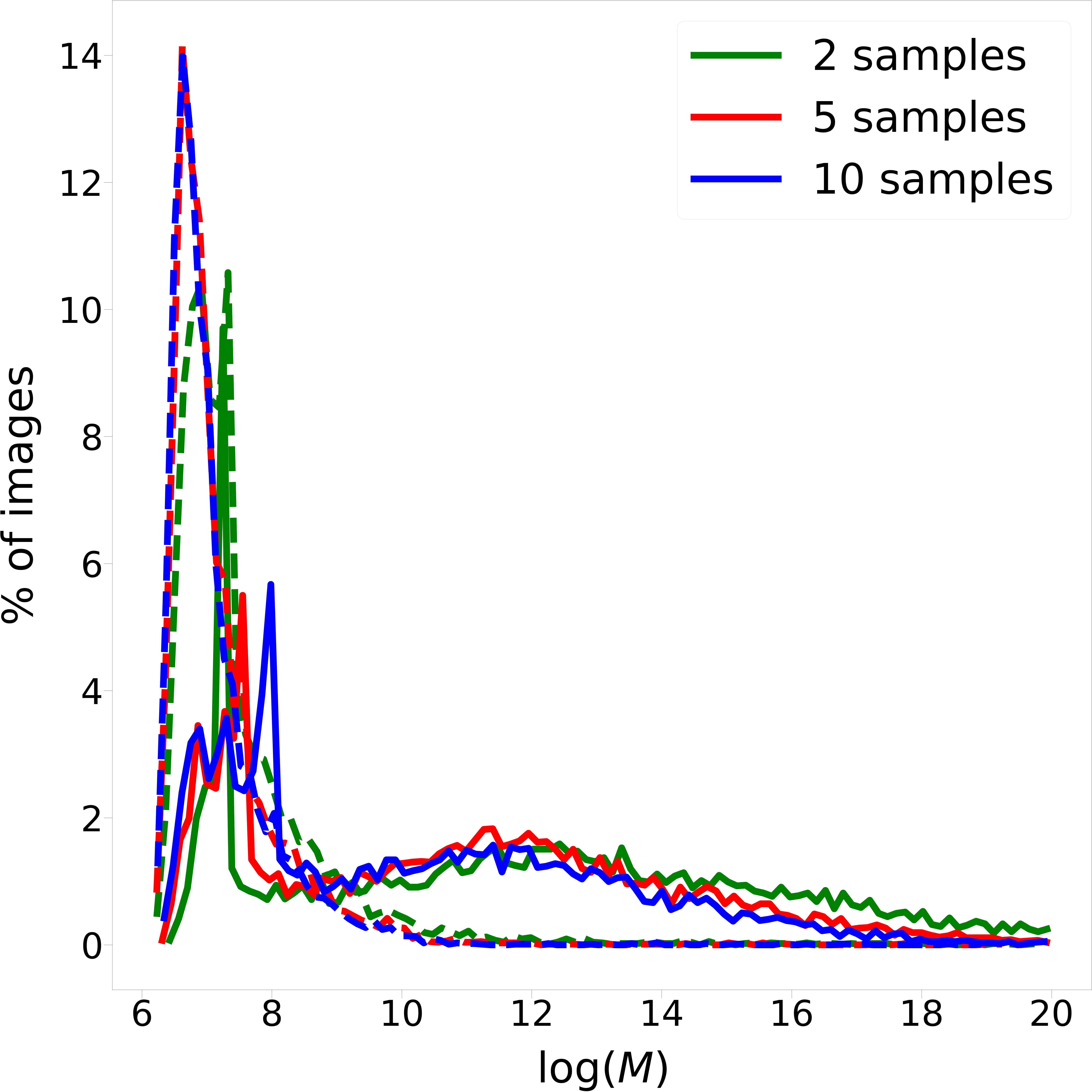}
\Description[The image illustrates the distribution of $\log{(M)}$ values for the ${\mathbb{C}_{1}}$ neural network architecture using the MNIST dataset as in-distribution (ID) data and Fashion-MNIST as out-of-distribution (OOD) data. The ID data distributions are represented by solid lines, while the OOD data distributions are shown with dashed lines. The plot includes various scenarios with different numbers of weight samples. Notably, the distributions for different numbers of samples taken almost overlap with each other, indicating that the change in the number of weight samples does not significantly affect the distinction between ID and OOD data in terms of the $\log{(M)}$ values.]{}
\caption{Distribution of $\log{(M)}$ values for ${\mathbb{C}_{1}}$ architecture with MNIST dataset as ID data (solid lines) and Fashion-MNIST as OOD data (dashed lines) for different number of weight samples. It can be seen that the distributions almost overlap with each other for different number of samples taken.}
\label{fig:numsample}
\end{figure}

\section{Discussion}
As mentioned in section 3, proposed probabilistic trust intervals can be interpreted as the space of directed noise. It is, therefore, important to discuss the scenario where instead of learning \(\sigma\), a trivial approach is adopted in which small random perturbations are added to the weights of an already trained neural network. We observed in our experiments that this trivial approach results in degradation of test accuracy if the strength of perturbations is too high and if it is too low then test accuracy is preserved but the performance on the task of OOD input detection is suboptimal. For example, when we directly added perturbations sampled from \(\mathcal{N}(0, 1)\) to the weights of ${\mathbb{C}_{1}}$, the test accuracy on MNIST dataset reduced from close to 98\% to \(\approx\) 10\% and when we sampled the perturbations from \(\mathcal{N}(0, 0.01)\), the FPR at 95\% TPR for Fashion-MNIST was \(\approx\) 55\% which is considerably poor as compared to what is observed for the corresponding PNN, shown in Table~\ref{tab:fpr}. In fact, handcrafting the strength of perturbations is nearly impossible because in complex architectures it is very difficult to predict the effect of perturbing weights on the outputs if done manually. Our proposed cost function automates this process and \(\sigma\) is learnt in such a way that both the requirements of OOD input detection and test accuracy are satisfied simultaneously. 

Further, it is also interesting to discuss the effect of number of siblings on OOD detection. For this we consider PNN corresponding to ${\mathbb{C}_{1}}$ trained on MNIST dataset. We sample 2, 5, and 10 values for weight parameters and evaluate the created sets of siblings for ID and OOD (Fashion-MNIST) samples. Similar to the experiments described in previous section, distributions of values of $M$ are plotted in Fig.~\ref{fig:numsample}. As we see, the distributions corresponding to OOD samples for different number of weight samples are almost overlapping each other. There is a little improvement with increasing number of weight samples in the performance on ID data. Overall the plots show that 2 samples is a reasonable choice to get the desired results while keeping the amount of computation during inference under control.

\section{Conclusion}

In this paper, we propose a simple approach to enable OOD detection on already trained deep neural network classifiers using the defined probabilistic trust intervals and optimized for ID data around each weight parameter. We also designed a measure of agreement for a robust detection using the resultant PNNs. Once trained PNN can be safely deployed with a constant memory and computation overhead. PNNs are able to distinguish adversarial and noisy inputs as well. We have used Gaussian distribution to model the probabilistic nature of the trust interval, in future other distributions can also be considered. Another possible future direction is to explore the possibility of removing redundant and very small sized intervals to reduce both, the time consumed in sampling and memory required to store interval parameters. PNNs can also be explored for uncertainty estimation in applications such as regression and segmentation.

\bibliographystyle{ACM-Reference-Format}
\bibliography{sample-base}

\end{document}